\documentclass[letterpaper, 10 pt, conference]{ieeeconf}  

\IEEEoverridecommandlockouts                              

\usepackage{graphics} 
\usepackage{epsfig} 
\usepackage{mathptmx} 
\usepackage{times} 
\usepackage{url}
\usepackage{amsmath} 
\usepackage{multirow}
\usepackage{amssymb}  
\usepackage{color}
\usepackage{gensymb}
\usepackage{siunitx}
\usepackage{lipsum}
\usepackage{mathtools, nccmath}
\usepackage{cuted}
\usepackage[space]{cite}

\newcommand{\lb}[1]{\underline{#1}}
\newcommand{\ub}[1]{\overline{#1}}

\title{\LARGE \bf
Trajectory Optimization for Manipulation of Deformable Objects:\\ Assembly of Belt Drive Units
}

\author{
  \authorblockN{Shiyu Jin$^{1}$, 
 Diego Romeres$^{2}$, Arvind Ragunathan$^{2}$, Devesh K. Jha$^{2}$ and Masayoshi Tomizuka$^{1}$} 
  \authorblockA{
     $^{1}$University of California, Berkeley, $^{2}$Mitsubishi Electric Research Laboratories\\
    {\tt\small $^{1}$<jsy,tomizuka>@berkeley.edu, $^{2}$<romeres,ragunathan,jha>@merl.com}} \

}

\begin{document}

\maketitle

\thispagestyle{empty}
\pagestyle{empty}


\begin{abstract}
This paper presents a novel trajectory optimization formulation to solve the robotic assembly of the belt drive unit. Robotic manipulations involving contacts and deformable objects are challenging in both dynamic modeling and trajectory planning. For modeling, variations in the belt tension and contact forces between the belt and the pulley could dramatically change the system dynamics. For trajectory planning, it is computationally expensive to plan trajectories for such hybrid dynamical systems as it usually requires planning for discrete modes separately.
In this work, we formulate the belt drive unit assembly task as a trajectory optimization problem with complementarity constraints to avoid explicitly imposing contact mode sequences. The problem is solved as a mathematical program with complementarity constraints (MPCC) to obtain feasible and efficient assembly trajectories. We validate the proposed method both in simulations with a physics engine and in real-world experiments with a robotic manipulator.  

\end{abstract}

\section{Introduction}\label{sec:introduction}
\begin{figure}
	\centering
    \includegraphics[scale=0.45]{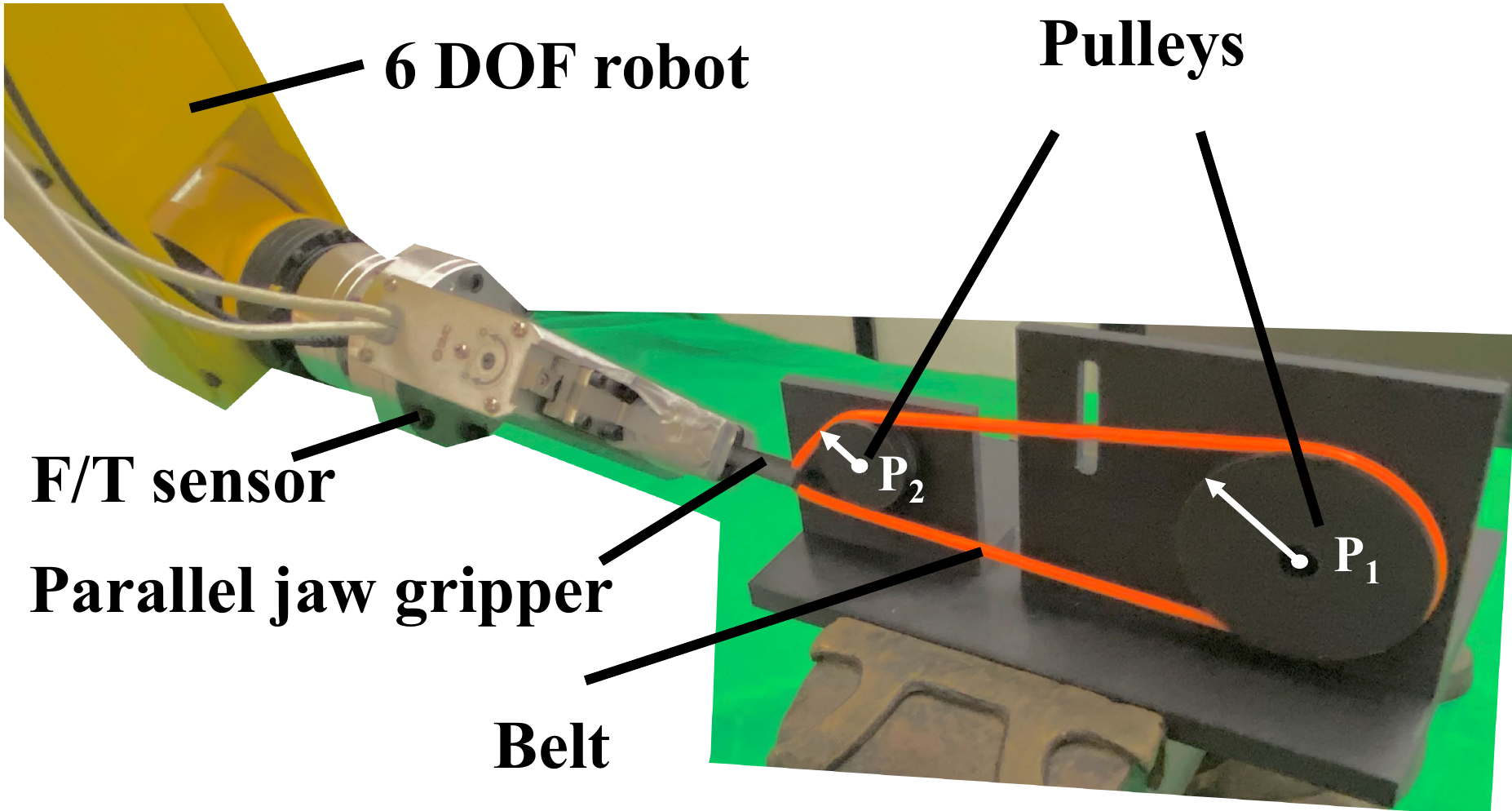}
    \caption{Belt drive unit assembly task. The robot grips a polyurethane belt and assembles it on two pulleys, $P_1$ and $P_2$.} 
    \label{fig:setup}
\end{figure}

While we have seen tremendous developments in the fields of artificial intelligence in recent years~\cite{lecun2015deep, krizhevsky2017imagenet, silver2017mastering,jin2021contact}, robots can achieve only limited autonomy during manipulation tasks~\cite{mason2018toward}. One of the biggest challenges that restricts general-purpose manipulation algorithms is contact dynamics. Contact-rich manipulation tasks are difficult to solve from both modeling and optimization perspectives.
The manipulation problems become further challenging when the manipulated objects are deformable. These kinds of objects are ubiquitous in a lot of assembly problems, and yet they remain poorly understood. 
In the assembly challenge competition in World Robot Summit 2018\footnote{\url{https://worldrobotsummit.org/en/about/}}, assembly of a polyurethane belt onto pulleys (see Figure~\ref{fig:setup}) was one of the most challenging tasks \cite{drigalski2019}. While there have been attempts to solve manipulation problems involving deformable objects \cite{Zheng1991,Ramirez_Alpizar2014,zhu2020,Tatemura2020,Tang2018,jinreal}, there is no general approach to it.

In this paper, we consider the problem of wrapping a belt around a two pulleys system, considering as use case the challenge introduced in the World Robot Summit 2018. Working with a deformable object like the belt presents several challenges. These include: (i) infinite degrees of freedom for the belt; (ii) contact rich manipulation; and (iii) long-horizon planning problem.

Optimization-based planning and control may be applied to various problems in robotic manipulation. Given a controlled dynamical system, $\dot{x} = f(x,u)$, trajectory optimization aims to design a finite-time input trajectory, $u(t), \forall t \in [0,T]$, which minimizes some cost functions over the resulting input and state trajectories \cite{posa2014, ourICRApaper20, yunt2006trajectory}. 
In the belt drive unit assembly, variations in the belt tensions and contact forces between the belt and the pulleys result in a hybrid dynamical system. Elastic force can exist or not, depending on whether the belt is slack or stretched. Contact forces can exist or not, depending on whether the belt contacts the pulley or not. Both elastic and contact forces might greatly impact the system dynamics. Planning for such a hybrid system usually requires planning for each dynamic system separately. There are many existing works on trajectory planning for hybrid systems \cite{Goebel2009,Fierro2001,Kaelbling2011}. But the drawback is that those methods require a task-specific mode schedule, which may bring extensive efforts in modeling and parameter tuning.

Inspired by the work on trajectory optimization of rigid bodies through contact \cite{posa2014,yunt2006trajectory, yunt2007combined,yunt2011augmented}, we model the physics of contacts and the elastic properties through complementarity constraints. The elastic belt is modeled through a 3D keypoint representation. The hybrid behavior of the keypoints is captured by the complementarity constraints. We formulate the trajectory optimization problem as a Mathematical Program with Complementarity Constraints (MPCC) \cite{mpcc2008}. We successfully solve the MPCC to compute feasible and efficient trajectories to assemble the belt drive unit. Finally, we implement the solution into the real system with a controller to track the optimized trajectory.

The main contributions presented in this work are:
\begin{enumerate}
    \item Trajectory optimization formulation for deformable objects manipulation with complementarity constraints. This provides a general-purpose, mathematical framework to tackle these problems.
    \item Introduction of $3$D keypoint representation for deformable objects.
    \item Validation of the proposed approach through simulation as well as real experiments.
    
\end{enumerate}


\section{Related Work}\label{sec:related_work}
Deformable linear (one-dimensional) object manipulation has been studied for decades. A randomized algorithm was proposed to plan a collision-free path for elastic objects \cite{Lamiraux2001}. Minimal-energy curves were applied to plan paths for deformable linear objects in stable configurations \cite{Moll2006}. In \cite{Navarro_Alarcon2013}, a local deformation model approximation method was proposed to control the soft objects to desired shapes. The authors of \cite{zhu2018,Jin2019} extended the local deformation model to the manipulation of cables. A deep neural network was trained to manipulate a rope to target shape based on a sequence of images \cite{Nair2017}. However, those works do not consider the interaction between the deformable cables and the environment. In~\cite{Zheng1991}, the authors proposed a strategy to assemble a flexible beam into a rigid hole. An optimization-based trajectory planning was utilized to assembly ring-shaped elastic objects in \cite{Ramirez_Alpizar2014}, but the authors only validated their method in simulation. In \cite{zhu2020}, the authors took the advantage of environmental contacts to shape deformable linear objects by a vision-based contact detector. The authors of \cite{Tatemura2020} considered a scenario to assemble the roller chain to sprockets. Their strategy successfully assemble the chain but lacks in generalization because each step is engineered for the specific system. To advance the research on robotic manipulation, the World Robot Summit 2018 proposed a competition on assembly challenges \cite{drigalski2019}. The challenge highlighted the complexity of solving manipulation tasks in a general manner, which still remains an open problem.

Optimization-based methods have been successfully implemented in many trajectory planning scenarios~\cite{chomp,stomp,trajopt}. \cite{posa2014} proposed a trajectory optimization method for rigid bodies contacting the environment. They formulated an MPCC to eliminate the prior mode ordering in discontinuous dynamics due to inelastic impacts and Coulomb friction. The MPCC framework was extended to a quadrotor with a cable-suspended payload system in \cite{Foehn2017}. The complementarity constraint was utilized to model the limitation of a non-stretchable cable length. Inspired by \cite{yunt2006trajectory, yunt2007combined,yunt2011augmented,posa2014,Foehn2017}, we introduce complementarity constraints to the belt drive unit assembly task to avoid the hybrid modes selection due to elastic force in the belt and contact force between the belt and the pulleys. We extend the keypoints representation introduced for rigid objects, \cite{mauelli2019}, to model elastic objects like the belt and to formulate an MPCC to perform the assembly.

\section{Problem Formulation}
\label{sec:problem_formulation}

\begin{figure}
	\centering
    \includegraphics[scale=0.25]{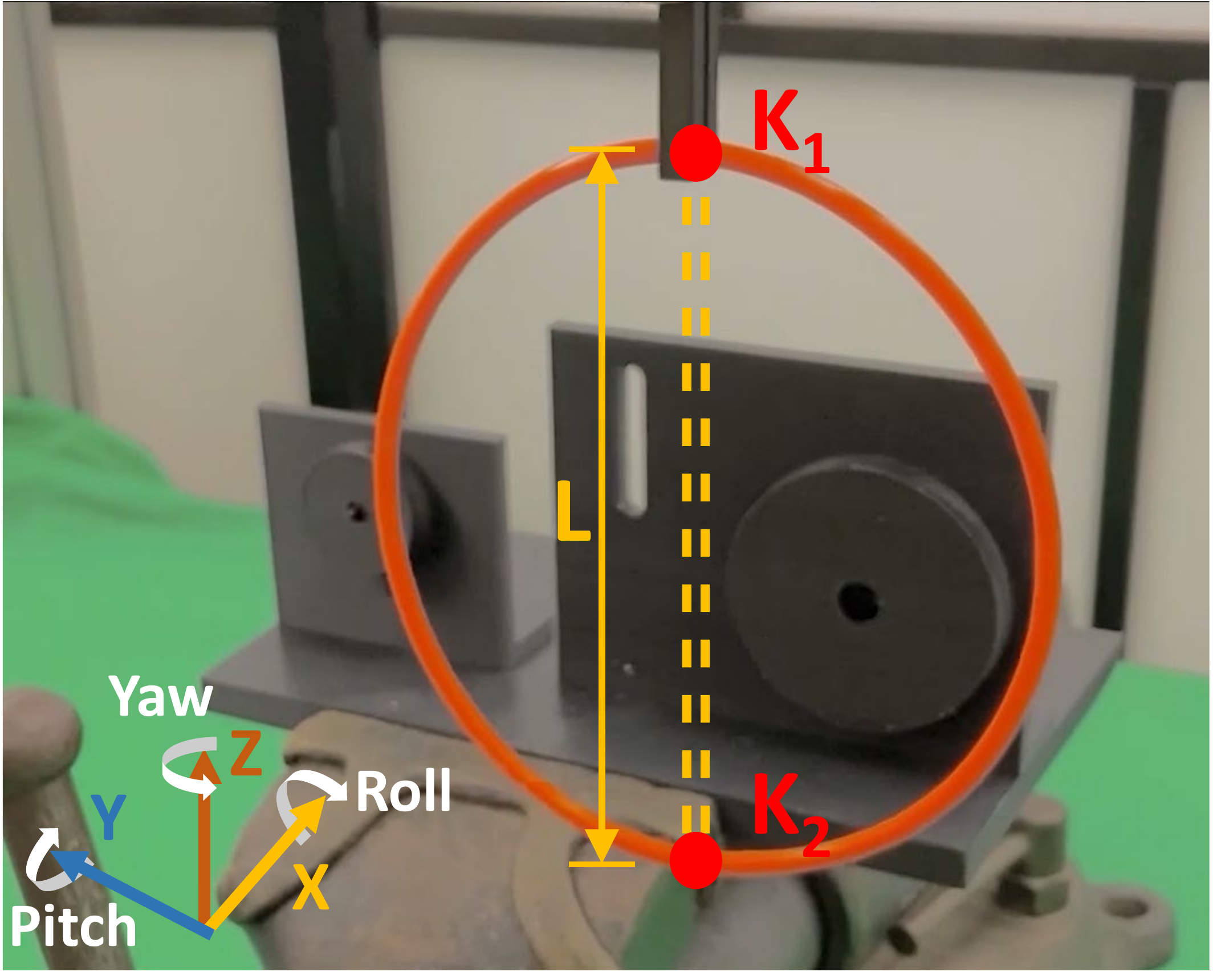}
    \caption{Initial belt configuration, $\rho_0$, with keypoints representation. The red dots represent the ``grasped" keypoint $K_1$ and ``opposite" keypoint $K_2$. The yellow dashed line shows the virtual cable, $\mathcal{C}$, of length $L$.}
    \label{fig:keypoints}
\end{figure}

The belt drive unit consists of two pulleys attached to a base and of a deformable and stretchable belt as shown in Figure~\ref{fig:setup}. The belt is assumed to be composed of a homogeneous isotropic linear elastic material which is a common assumption in mechanics. The pulleys have known geometries and can rotate freely around the shafts axis. The base of the belt drive unit is fixed to the workbench in a known pose. We assume that at the initial configuration, called $\rho_0$, the belt is grasped and lifted by a gripper held by a robotic manipulator, and the belt is freely hanging under the effect of gravity, see Figure~\ref{fig:keypoints}. The task objective is to wrap the belt around the two pulleys as shown in Figure~\ref{fig:setup}.

Inspired by recent work \cite{mauelli2019}, we introduce a 3D keypoints representation for deformable objects. This representation consists of identifying points in the object that are representative of the whole object. 
With the proposed representation, the problem is mathematically tractable with a finite, low dimensional state space and interpretable constraints and cost function. In particular, we select two 3D keypoints for the belt as shown in Figure~\ref{fig:keypoints}. 
The ``grasped" keypoint, $K_1$, corresponds to the point-mass on the belt grasped by the robot gripper, and the ``opposite" keypoint, $K_2$, which is the point on the belt that is the furthest away from $K_1$ when the belt is in configuration $\rho_0$. In the proposed keypoint representation, configuration $\rho_0$ can be represented by a virtual elastic cable, $\mathcal{C}$, that connects $K_1$ and $K_2$. The generalized coordinates of the system can now be described as $q = [K_1^x, K_1^y,K_1^z, K_2^x, K_2^y,K_2^z,K_1^{roll}, K_1^{pitch},K_1^{yaw}]^\top \in \mathbb{R}^9$, where $K_1^x, K_1^y,K_1^z, K_2^x, K_2^y,K_2^z$ are the Cartesian coordinates of two keypoints and $K_1^{roll}, K_1^{pitch},K_1^{yaw}$ are the orientation of $K_1$ with reference frame shown in Figure~\ref{fig:keypoints}. We utilize the orientation of $K_1$ to express the rotation and the twist of the belt. The action space $u=[F_x, F_y, F_z, M_x, M_y, M_z]^\top \in \mathbb{R}^6$ is the vector of forces and torques that are applied to $K_1$ through the gripper. This makes the belt drive unit an underactuated system as we cannot directly control $K_2$. Finally, we assume that in configuration $\rho_0$ the ellipsoidal shape of the belt is large enough to go around the first pulley~$P_1$.

%

\subsection{Subtasks Decomposition} \label{subsec:subtasks}
\begin{figure}
	\centering
    \includegraphics[scale=0.25]{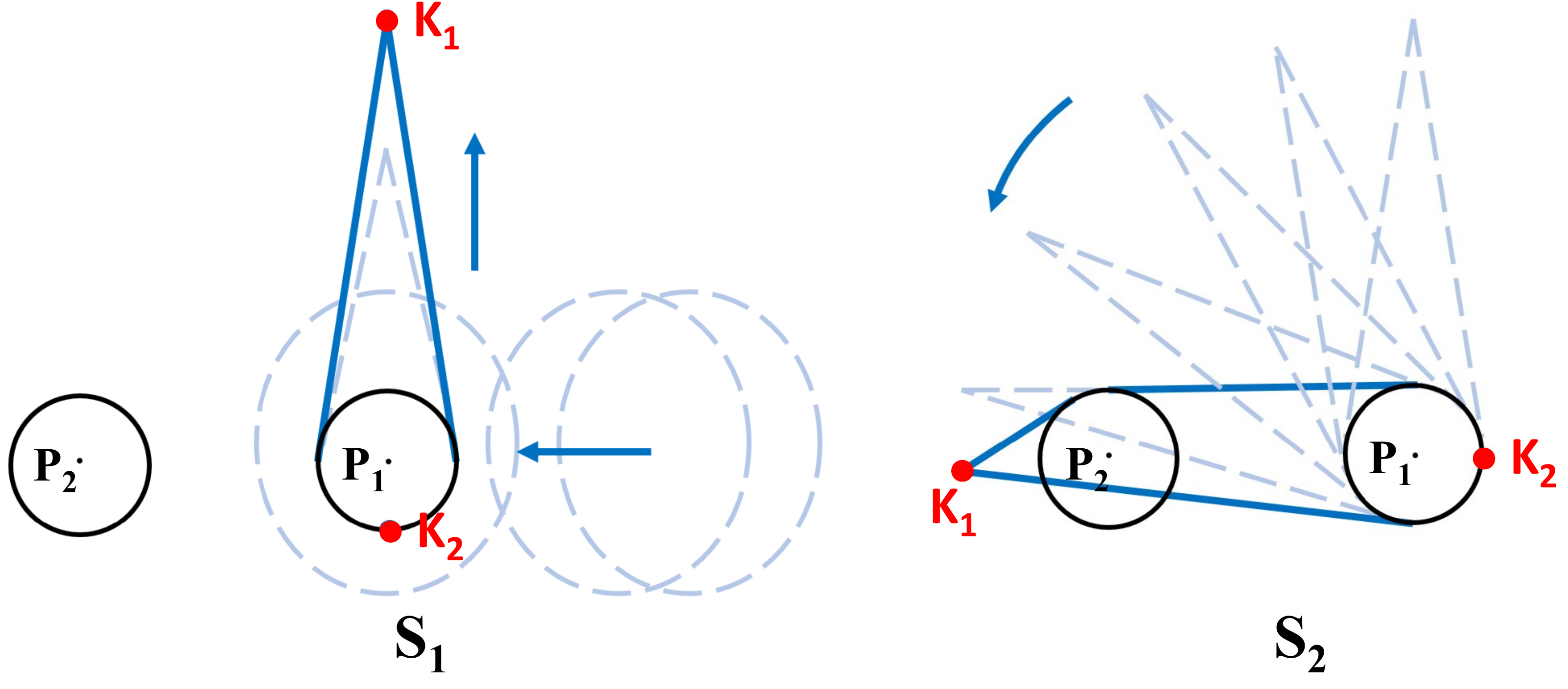}
    \caption{Two subtasks decomposition. $P_1$ and $P_2$ are two pulleys. The blue lines represent the belt gripped at keypoint $K_1$ by a robot. $S_1$: The belt wraps the first pulley $P_1$ and it stretched. $S_2$: The belt rotates around the first pulley and is assembled onto the second pulley $P_2$.} 
    \label{fig:subtasks}
\end{figure}

Belt drive unit assembly is a complex task that requires a long-horizon planner. As often proposed in the literature, long-horizon planning tasks are decomposed into subtasks to reduce complexity. The belt drive unit assembly can have highly engineered solutions with a dense sequence of subtasks and simple planners whose subgoals are trivial to reach. 
However, this kind of approach requires extensive effort in parameter tuning and engineering work and lacks generality, since the goals of the subtasks need to be redefined as the scenario changes. We partially address this problem by reducing the number of subtasks to two. Following a logic similar to a human's approach, the first subtask, $S_1$, corresponds to wrap the belt around the first pulley, and the second subtask, $S_2$, corresponds to wrap the second pulley keeping the belt taut to maintain the wrap around the first pulley, see Figure~\ref{fig:subtasks}. In a qualitative description, in $S_1$, the belt has to avoid the outer surface of the first pulley $P_1$ and $K_2$ has to get into the groove creating a contact force, while $K_1$ is stretched until the belt is taut. In $S_2$, the belt is assembled on the second pulley $P_2$ by rotation around the first pulley. During the rotation, the belt should remain taut in order to remain in the groove of the first pulley. Finally, the belt has to hook the internal groove of $P_2$ and $K_1$ has to reach the bottom of the second pulley to accomplish the task. 

Given the proposed 3D keypoint representation of the belt drive unit, we can formulate a trajectory optimization problem, that uses complementarity constraints to model the contacts and the deformation of the belt, to solve each of the two subtasks. The two optimized trajectories are then executed sequentially in order to accomplish the task, the final condition of $S_1$ corresponds to the initial condition of~$S_2$.

\section{Trajectory Optimization for the Belt Drive Unit Assembly}
\label{sec:trajOpt}

We approach the belt drive unit assembly as a trajectory optimization problem formulated as an MPCC. The complexity of this problem is given by the presence of hybrid nonlinear dynamics due to contacts that may happen between the pulleys and the belt, the elastic properties of the belt, the obstacle avoidance constraints, and the long planning horizon. A trajectory optimization problem is solved for each of the two subtasks described in Sec.~\ref{subsec:subtasks} of the form
\begin{subequations}
\begin{align}
    \min\limits_{q, \dot{q},u, \lambda}  \quad      &\, L(q, \dot{q},u, \lambda) \label{opt:obj} \\
    \text{s.t.} \quad &\, H(q)\ddot{q} + C(q,\dot{q}) + G(q) = B(q)u+\lambda \label{opt:dynamic} \\
                &\, g(q, \dot{q},u, \lambda) \leq 0 \label{opt:ineq} \\
                &\, \lb{q} \leq q \leq \ub{q}, \, \lb{\dot{q}} \leq \dot{q} \leq \ub{\dot{q}}, \, \lb{u} \leq u \leq \ub{u}, \, \lb{\lambda} \leq \lambda \leq \ub{\lambda} \label{opt:bnds}
\end{align}\label{opt}
\end{subequations}
where $L(q, \dot{q},u, \lambda)$ is the cost function, $q \in \mathbb{R}^9$ are the generalized coordinates described in Sec.~\ref{sec:problem_formulation}, $\dot{q}$ and $\ddot{q}$ are its first and second order time derivatives, $u \in \mathbb{R}^6$ is the control input and $\lambda$ are the external forces acting on the belt. Eq.~\eqref{opt:dynamic} is the forward dynamics, where $H(q), C(q,\dot{q}), G(q)$ are the inertial matrix, the Coriolis terms, and the gravitational forces, respectively. $B(q)$ is input mapping. The general nonlinear constraints~\eqref{opt:ineq} include the complementarity constraints and collision avoidance. Eq.~\eqref{opt:bnds} represents the lower and upper bounds of the optimization variables.

We solve \eqref{opt} as a nonlinear program and use a direct approach which in general has better numerical properties than shooting methods, and we can exploit the sparsity structure of the problem. We directly optimize the feasible general coordinates and its first-time derivative, the control inputs, and the external forces. The discretization of the forward dynamics is obtained by the trapezoidal rule. 
The formulation of the contact and elastic forces as complementarity constraints fits naturally well in this formulation. In practice, for numerical advantages, we use a relaxed version of the complementarity constraints as described in~\cite{raghunathan2005interior}.

In the following, the dynamical constraints and the cost function for MPCC \eqref{opt} are described for the two subtasks.




\subsection{Dynamics Constraints}

\begin{figure}
	\centering
    \includegraphics[scale=0.4]{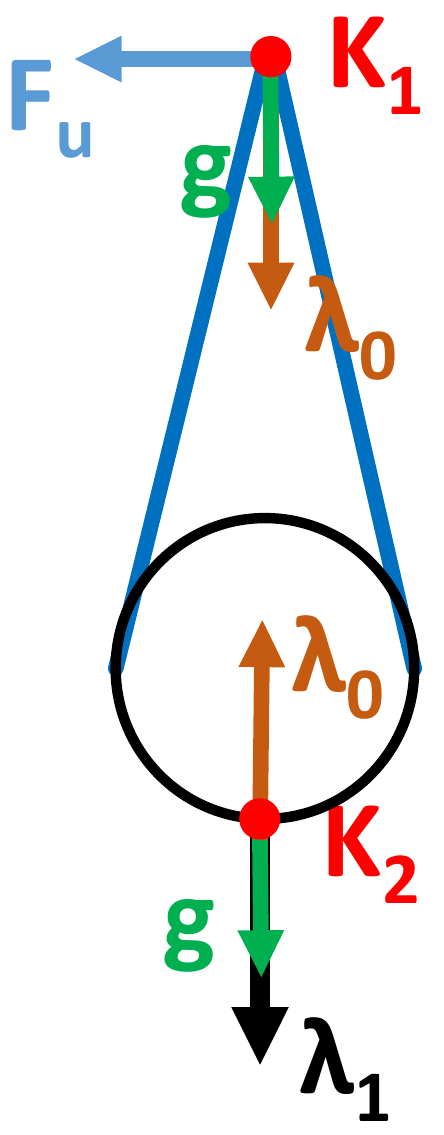}
    \caption{Force analysis at the two keypoints. $F_u$ is the control input force. $\lambda_0$ is the elastic force. $\lambda_1$ is the contact normal force. $g$ is the gravity force.}
    \label{fig:forces}
\end{figure}

The system is composed of the two keypoints, $K_1$ and $K_2$, and the virtual elastic cable, $\mathcal{C}$. It is modeled similarly to a mechanical mass-spring-damper second-order system, with an actuator acting on $K_1$ and subject to the gravity and the external forces given by the elastic force $\lambda_0$ and the normal force $\lambda_1$ experienced during contacts. Figure~\ref{fig:forces} shows a schematic example of the forces that act on the system at the end of subtask $S_1$. 
%
The system dynamics are defined as 
\begin{align}
    \dot{x} & = Ax + Bu + G + f(x,\lambda) 
\end{align}
where $x=[q^\top, \dot{q}^\top]^\top \in \mathbb{R}^{18}$ is the system state and $\lambda = [\lambda_0^\top, \lambda_1^\top]^\top$ is the vector of the external forces. The state transition matrix
 $$A=\begin{bmatrix}
0_{9\times 9} & I_{9\times 9}  \\
0_{3\times 9} & -\frac{k_d}{m_1}I_{3\times  3}   \quad \frac{k_d}{m_1} I_{3\times  3} \quad 0_{3\times 3}\\
0_{3\times 9} & \frac{k_d}{m_2}I_{3\times  3}  \quad -\frac{k_d}{m_2}I_{3\times  3} \quad 0_{3\times 3}\\
0_{3\times 9} & 0_{3\times 9}  \\
\end{bmatrix} \in \mathbb{R}^{18 \times 18}$$ 
represents the effect of the mass-spring-damper system with $k_d$ the damping coefficient and $m_1$, $m_2$ are the masses of the keypoints $K_1$ and $K_2$, respectively. The input matrix
$$B=\begin{bmatrix}
0_{9\times 6}  \\
\frac{I_{3\times  3}}{m_1} \quad 0_{3\times 3}\\
0_{3\times 6}\\
0_{3\times 3} \quad \frac{I_{3\times  3}}{M_1}
\end{bmatrix} \in \mathbb{R}^{18 \times 6}$$
maps the 6-dimensional end-effector force/torque input $u$ to the linear and angular acceleration of the ``grasped" keypoint $K_1$. $M_1$ is the moment of inertia of $K_1$. The gravitational acceleration is applied to the two keypoints through the vector 
$G=\begin{bmatrix}
0_{1\times 11}, \, -g, \, 0_{1\times 2}, \, -g, \, 0_{1\times 3} 
\end{bmatrix}^\top \in \mathbb{R}^{18 \times 1}$.

The contribution of the external forces is now given by the sum of the elastic and normal force $f(x,\lambda)=\lambda_0+\lambda_1$. The elastic force is defined as
$$\lambda_0=\begin{bmatrix}
0_{3\times 9}, \,
-\frac{I_{3\times  3}}{m_1}, \,
\frac{I_{3\times 3}}{m_2},
\, 0_{3\times 3}
\end{bmatrix}^\top 
\Pi_{K_1, p} \,
\bar{\lambda}_0 \in \mathbb{R}^{18 \times 1}$$
where $\bar{\lambda}_0 \in \mathbb{R}$ is the magnitude of the elastic force and is the variable optimized, $\Pi_{K_1, p} = \frac{[(K_1^x - p^x),(K_1^y - p^y),(K_1^z - p^z)]^\top}{\lvert \lvert K_1 - p \rvert \rvert}$ is the projection operator of the elastic force into the 3 axis. The point $p$ is $K_2$ in $S_1$ and $O_1$ in $S_2$ for simplicity of computation. $O_1$ is the position of the first pulley's center.
The normal force due to the contacts between the pulley and the keypoint $K_2$ is defined as
$$ \lambda_1 = \begin{bmatrix}
0_{3\times 12}, \,
-\frac{I_{3\times  3}}{m_2}, \,
0_{3 \times 3}
\end{bmatrix}^\top \Pi_{O_1, K_2}  \bar{\lambda}_1 \in \mathbb{R}^{18 \times 1}$$ 
where $\bar{\lambda}_1 \in \mathbb{R}$ is the magnitude of the normal force and is the variable optimized and $\Pi_{O_1, K_2} = \frac{[(o_1^x - K_2^x),(o_1^y - K_2^y),(o_1^z - K_2^z)]^\top} 
{\lvert \lvert o_1 - K_2 \rvert \rvert}$ is the projection operator of the normal force into the 3-axis.



\subsection{Complementarity Constraints} \label{subsec:CC}
In order to model the hybrid dynamics due to elastic force and contact force, we use complementarity constraints
\begin{align}
    0 \leq g(\cdot) \quad \perp \quad h(\cdot) \geq 0
\end{align}
Complementary constraints are a way to model constraints that are combinatorial in nature and impose the positivity and orthogonality of the variables. 

\textbf{Elastic force constraint.} The first complementarity constraint is formulated as
\begin{subequations}
\begin{align}
    \lambda_2 = \frac{\bar{\lambda}_0}{k_p} + L - l(x) & \geq 0 \label{eq:elasticity}\\
   \bar{ \lambda}_0 & \geq 0\\
    \bar{\lambda}_0 \lambda_2 & = 0
\end{align}
\end{subequations}
where $L$ and $k_p$ are respectively the length at configuration $\rho_0$ and the stiffness coefficient of the virtual elastic cable,~$\mathcal{C}$. The length of $\mathcal{C}$ at each temporal instant is $l(x) = \lvert \lvert K_1 - K_2 \rvert \rvert$ in $S_1$, and $l(x) = \lvert \lvert K_1 - O_1 \rvert \rvert + r_1$ in $S_2$, where $r_1$ denotes the radius of $P_1$. The pulley center $O_1$ is chosen because it is a fixed known point while the pulley is rotating. From eq.~\eqref{eq:elasticity} the elasticity of the belt is defined as proportional to the length $L - l(x)$ and depends on the stiffness coefficient $k_p$. $\lambda_2$ is an algebraic variable. If the cable is stretched, then $L < l(x)$, $\bar{\lambda}_0>0$, and $\lambda_2=0$. If the cable is slack, then $L > l(x)$, $\bar{\lambda}_0=0$, and $\lambda_2>0$.

\textbf{Contact force constraint.} The second complementarity constraint is formulated as
\begin{subequations}
\begin{align}
    \lambda_3 =\sqrt{\lvert \lvert K_2 - O_e \rvert \rvert^2 + \epsilon^2} & \geq \epsilon\\
    \bar{\lambda}_1 & \geq 0\\
    \bar{\lambda}_1 \lambda_3 & = 0
\end{align}
\end{subequations}
where $O_e$ is the contact point on the edge of $P_1$. $\epsilon$ denotes a small number to relax the complementarity constraint~\cite{raghunathan2005interior}. $\lambda_3$ is the algebraic variable describes whether the belt contacts the pulley. If the belt contacts the pulley, then $\lambda_3 = \epsilon$, and contact force $\bar{\lambda}_1 \geq 0$. If there is no contact, then $\lambda_3 >\epsilon$, and contact force $\bar{\lambda}_1 = 0$.

\subsection{Obstacle avoidance}
This constraint imposes that the keypoints cannot penetrate into the pulleys. Each pulley is approximated with an ellipsoid, since there is a known analytical expression of the distance function between a point and an ellipsoid. The obstacle avoidance constraints between a keypoint $K_i$ and a pulley $P_j$ can be denoted as $distance(K_i, P_j) = \sqrt{(K_i - O_j)^\top S(K_i - O_j) } - 1 \geq 0$, where $S = diag\{1/{a^2}, 1/{b^2}, 1/{c^2}\}$ is a diagonal matrix, $a,b,c$ are half the length of the principal axes. $O_j$ denotes the center of pulley $P_j$. 

\subsection{Physics Limitation}
The belt might break if stretched over a certain limit, this condition is approximated by constraining the length of the virtual cable $\mathcal{C}$, $l(x) \leq L_{max}$. Moreover, $L_{max}$ is assumed large enough for the loop to go around two pulleys.


\subsection{Cost Function}

We use a common quadratic cost function that penalizes the difference to the goal state $x^{goal}$ and the control input~$u(k)$:
\begin{equation}
\label{eq:cost_function}
\begin{aligned}
  J(x,u,\lambda) = \sum_{k=0}^{N} & (x(k)-x^{goal})^\top Q(x(k)-x^{goal}) +\\
  &u(k)^TRu(k) + w(\bar{\lambda}_0(k) - \bar{\lambda}_0^{desired})^2
\end{aligned}
\end{equation}
where the weights $Q$ and $R$ are diagonal matrices and $w$ is a scalar. Moreover, the term $w(\bar{\lambda}_0(k) - \bar{\lambda}_0^{desired})^2$ adds a soft constraint in the elastic force. A positive $\bar{\lambda}_0^{desired}$ encourages a solution with the belt in tension. This constraint is used in subtask $S_2$ to maintain the belt taut. Instead, in $S_1$ we set $w=0$.

\section{Experimental Results}
\label{results}
In this section, we present the results of the proposed method both in simulation using the physic engine MuJoCo~\cite{mujoco} and in a real system with a $6$-DoF manipulator. We use the Ipopt \cite{ipopt} solver in a python wrapper. 


\subsection{Simulations}
\subsubsection{Simulation Setup}
The belt drive unit is represented in a simulated environment in MuJoCo as shown in the top left corner of Fig.~\ref{fig:scenarios}. The environment includes two pulleys and one belt. The radius of the pulleys are of $30[mm]$ for $P_1$ and $15[mm]$ for $P_2$. 
The belt is composed of 41 linked objects called capsules in MuJoCo. Any two adjacent capsules are connected by two hinge joints and one prismatic joint. The physical properties of the simulated belt are tuned to resemble the belt of the real belt drive unit. The belt is held by a parallel gripper attached to a 6 DOF Fanuc LR-Mate 200iD. The purpose of the manipulator is to actuate the end-effector in order to track the optimal trajectory, but in simulation could be removed. 


\subsubsection{Trajectory Optimization in Different Scenarios}
\begin{figure}[h]
	\centering
    \includegraphics[scale=0.45]{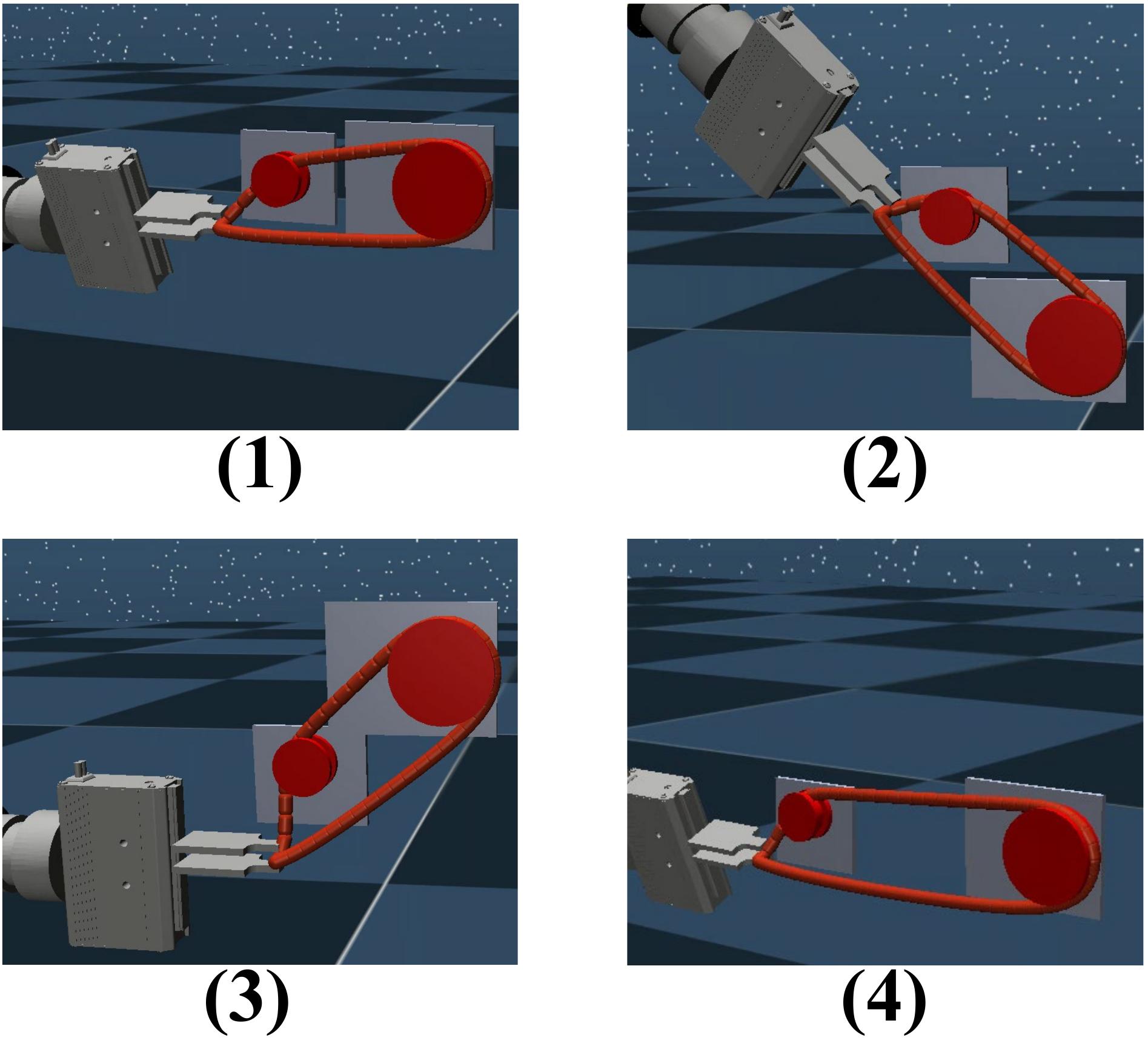}
    \caption{Snapshots of $4$ different simulation scenarios at the goal position of subtask $S_2$. The relative positions of the two pulleys vary in each scenario. 
    } 
    \label{fig:scenarios}
\end{figure}

In order to verify the generality of the approach to different known geometries, we consider 4 different scenarios, where the position of $P_1$ is fixed and the position of the smaller pulley $P_2$ varies, see Figure~\ref{fig:scenarios} and Table~\ref{table1}. Each pulley is modeled as three adjacent cylinders, and the two outer cylinders have larger radius than the inner one. The belts' lengths, $P_{belt}$, are chosen in each scenario based on the distance between two pulleys. The mass of the keypoints is $m_1 = m_2=0.042 [kg]$. The moment of inertia of $K_1$ is $M_1 = 10^{-7}[kgm^2]$. The belt's stiffness and damping coefficient are $k_p=63.34 [N/m]$ and $k_d=4.65 [Ns/m]$, respectively.

In the trajectory optimization formulation described in Section~\ref{sec:trajOpt}, the goal for subtask $S_1$ is set vertically above the pulley $P_1$ for keypoint $K_1$, and right under the pulley $P_1$ for keypoint $K_2$, respectively, e.g., $q_1^{goal} =~[0.10, 0.55, 0.53, 0.10, 0.23,0.34, 0, 0, 0]^T$ and both with zero velocity. In this substask there is a change of mode in the dynamics from no contact to contact between the belt and the environment and the deformation of the belt for reaching the desired target. 

In the second subtask $S_2$, the goal, $q_2^{goal}$, is set only for $K_1$ in both the Cartesian coordinates and angular orientation, according to the position of the pulley, $P_2$. A qualitative representation of the goal position is shown in Figure~\ref{fig:scenarios} for each of the scenarios. The desired $[K_1^{roll}, K_1^{pitch},K_1^{yaw}]^\top$ is $[-\pi/2, 0, \pi/2]^\top$. The twist of the virtual cable $\mathcal{C}$ approximates the twist of the belt which leads to the assemble onto the groove of the pulley $P_2$. Based on $q_2^{goal}$ and $k_p$ it is possible to compute the target elastic force as $\bar{\lambda}_0^{desired}$, which encourages the belt to be stretched during rotation.   

\begin{table*}
\centering
\caption{Simulation results in 4 scenarios. In each scenario the position of the pulley center $O_2$ varies.}
\label{table1}
\begin{tabular}{|l|l|c|c|c|c|c|}
\hline 
 &$P_{belt}$& $O_1$ (center of $P_1$) & $O_2$ (center of $P_2$) & Feasible trajectory & Successful assembly\\\hline 
 Scenario 1&$0.4m$&[0.100, 0.550, 0.340]&[0.100, 0.680, 0.340]&10/10&10/10\\\hline 
 Scenario 2&$0.4m$&[0.100, 0.550, 0.340]&[0.100, 0.642, 0.432]&10/10&8/10\\\hline 
 Scenario 3&$0.4m$&[0.100, 0.550, 0.340]&[0.100, 0.645,  0.275]&10/10&7/10\\\hline 
 Scenario 4&$0.6m$&[0.100, 0.550, 0.340]&[0.100, 0.780,  0.340]&10/10&9/10\\\hline 
\end{tabular}
\label{table1}
\end{table*}


We perform 10 experiments for each scenario. In each experiment, the goal positions of the ``grasped" keypoint $K_1$ are sampled from the normal distributions $\mathcal{N}(\mu_1,\,\Sigma)$ in $S_1$ and  $\mathcal{N}(\mu_2,\,\Sigma)$ in $S_2$. Where, $\mu_1, \, \mu_2 \in \mathbb{R}^3$ are the components $[K_1^x, K_1^y,K_1^z]^\top$ in a pre-selected successful $q_1^{goal}$ and $q_2^{goal}$ and $\Sigma~=~diag\{0.005,0.005,0.005\}$. The lower and upper constraints of position, velocity, tilt angle, and force are $\pm 1 m$, $\pm 0.5 m/s$, $\pm \pi$, and $\pm 50 N$, respectively.


\subsubsection{Results}
\begin{figure}
	\centering
    \includegraphics[scale=0.78]{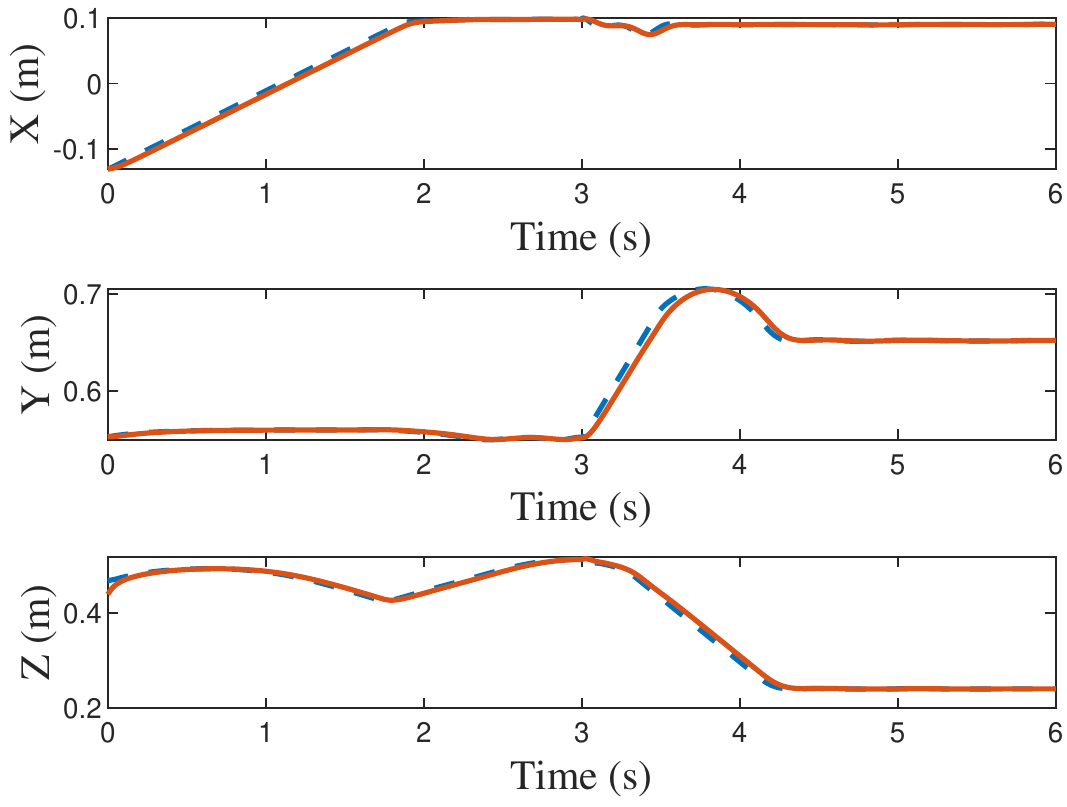}
    \caption{Trajectory of keypoint $K_1$ in a successful assembly for scenario Figure~\ref{fig:scenarios}(3). The dashed line is the reference trajectory obtained from MPCC. The solid line is the measured trajectory.}
    \label{fig:traj_control}
\end{figure}

The simulation results are shown in Table~\ref{table1}. We initialize the trajectory with all states $x(k)=x(0)$, where $k =0,1,..,N$. The solver finds a feasible trajectory for both subtasks given any sampled goals. The optimal trajectory obtained for $K_1$ is then tracked by the end-effector, and the assembly is completed successfully in $34/40$ experiments. The failure cases happen when the goal is sampled away from $q_1^{goal}$ or $q_2^{goal}$ because the belt fails to wrap around the pulley. The purpose of these experiments is to show that the engineering effort in finding the goal position for the subtask is reduced as it is not required to provide one specific point. But also the trade-off of having only two keypoints, more keypoints would make a more accurate model but also a more complex optimization problem. We use an Intel 12 Cores i7-9850H CPU @ 2.60GHz. The average computational time for one trajectory with $600$ time steps is $36.138\pm5.747[s]$. The computational time highly depends on the number of time steps selected. Figure~\ref{fig:traj_control} shows one full successful assembly trajectory for scenario Figure~\ref{fig:scenarios}(3). 

\subsection{Real-World Experiments}
\begin{figure}
	\centering
    \includegraphics[scale=0.26]{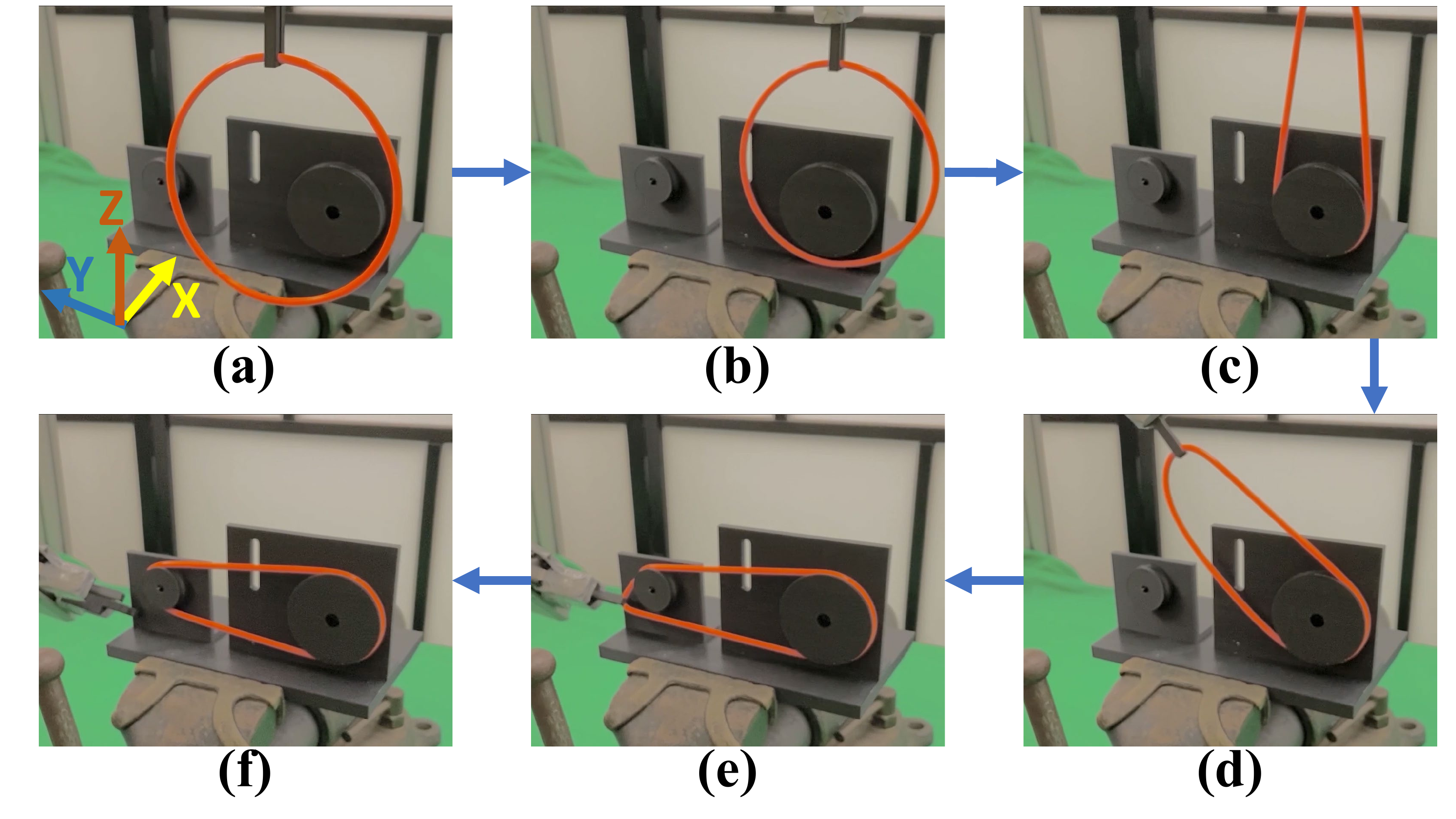}
    \caption{Snapshots of the experiment.}
    \label{fig:experiment}
\end{figure}
\subsubsection{Experimental Setup}
As shown in Figure~\ref{fig:setup}, the experiment environment includes a 6 DOF FANUC LR-Mate 200iD, an ATI Mini45 F/T sensor, and a 3D printed belt drive unit of the same dimensions in the assembly challenge \cite{drigalski2019} fixed on a vise. The belt is the same as in the challenge with length $0.40[m]$ and is gripped by a parallel jaw gripper. We assume no slip between the belt and the robot gripper. The pose of the pulleys is known exactly.

\subsubsection{Results}
\begin{figure}
	\centering
    \includegraphics[scale=0.45]{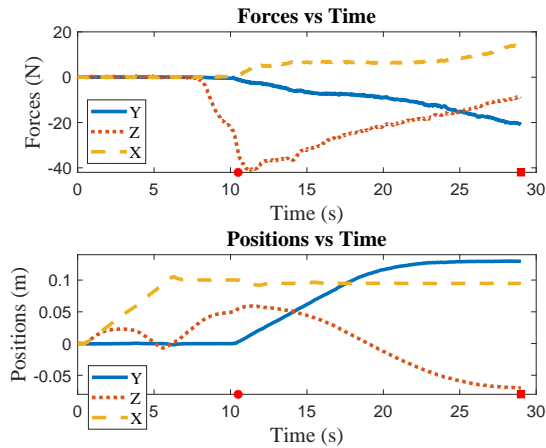}
    \caption{Forces and positions of the end-effector in a successful experiment. The red circle and square represent the end of subtask $S_1$, $S_2$,~respectively.}
    \label{fig:forceplot}
\end{figure}
Figure~\ref{fig:experiment} provides the snapshots of the main phases during the execution of a successful experiment. Figure~\ref{fig:forceplot} shows the trajectory of the gripper tip that corresponds to $K_1$ and the measured forces at the robot's wrist along the trajectory. In the beginning, (Figure~\ref{fig:experiment}a), the belt approaches the pulley and position $X$ increases and the forces are zero. The position $Z$ goes down at $5.57[s]$ to avoid the outer cylinder of the first pulley. At $6.29[s]$, position $X$ stops increasing because the pulley is reached (Figure~\ref{fig:experiment}b). Then the $Z$ position increases as the belt is lifted and contacts the pulley at $7.82[s]$ with a corresponding increase in force along the negative direction in $Z$. At $10.50[s]$, the system accomplishes $S_1$ (Figure \ref{fig:experiment}c). 
After that, the belt is rotated around $O_1$ (the $Z$ position decreases, and $Y$ position increases) while being stretched (Figure \ref{fig:experiment}d). In this phase, the measured net force is closed to the desired elastic force $\bar{\lambda}_0^{desired}$. The target orientations are $[K_1^{roll}, K_1^{pitch},K_1^{yaw}]^\top=[-\pi/2, 0, \pi/4]^\top$.  The belt is twisted so that it hooks the second pulley without jamming. Finally, the goal of subtask $S_2$ is reached at $29.00[s]$ (Figure~\ref{fig:experiment}e) and the gripper releases the belt (Figure~\ref{fig:experiment}f). 

The experiment has been repeated multiple times but given the robot's accuracy the results were similar to each other.

\section{Conclusion}
\label{sec:conclusion}

In this paper, we propose a trajectory optimization formulation to assemble the belt drive unit. We propose a 3D keypoints representation to model the elastic belt, which simplifies the complexity of the trajectory optimization problem. The problem is formulated as an MPCC with complementarity constraints to model the hybrid dynamics due to contact and elastic forces. Simulations results show that the proposed approach can find feasible trajectories for the belt drive unit assembly with known but variable geometry. To the best of our knowledge, this is the first work that formalizes the trajectory optimization problem for the belt drive unit assembly, and the solution works in the real system. Several future works are possible. The current method is based on the execution of an open-loop trajectory which could fail under uncertainties in the position of the pulleys or of the belt. Adding a feedback controller is fundamental for a more robust and reliable solution. Moreover, in order to improve the generality of the problem, we are interested in an autonomous selection of the 3D keypoints for a given task. Our formulation of a trajectory optimization problem for deformable objects using complementarity constraints is not limited to belt drive unit assembly. The proposed method might be applied to a wider range of tasks such as cable routing and wire harness. 
\newpage








\bibliographystyle{IEEEtran}
\bibliography{Ref}

\end{document}